\pdfoutput=1
\documentclass{article}
\usepackage{spconf,amsmath,graphicx}
\usepackage{multirow}
\usepackage{makecell}

\title{Multi-head Monotonic Chunkwise Attention for Online Speech Recognition}
\name{Baiji Liu$^1$, Songjun Cao$^2$, Sining Sun$^2$, Weibin Zhang$^{1}$, Long Ma$^2$\thanks{This work is supported in part by the National Natural Science Founding of China (61601187 ).}}
\address{ \large
\hskip -7pt 
\begin{tabular}{c}
$^1$South China University of Technology, GuangZhou, China \\
$^2$Tencent, China
\end{tabular}
}
\begin{document}
%
\maketitle
\begin{abstract}
The attention mechanism of the Listen, Attend and Spell (LAS) model requires the whole input sequence to calculate the attention context and thus is not suitable for online speech recognition. To deal with this problem, we propose multi-head monotonic chunk-wise attention (MTH-MoChA), an improved version of MoChA. MTH-MoChA splits the input sequence into small chunks and computes multi-head attentions over the chunks. We also explore useful training strategies such as LSTM pooling, minimum world error rate training and SpecAugment to further improve the performance of MTH-MoChA.  Experiments on \textit{AISHELL-1} data show that the proposed model, along with the training strategies, improve the character error rate (CER) of MoChA from 8.96\% to 7.68\% on test set. On another 18000 hours in-car speech data set, MTH-MoChA obtains 7.28\% CER, which is significantly better than a state-of-the-art hybrid system.
\end{abstract}
\section{Introduction}
\label{sec:intro}

A conventional hybrid automatic speech recognition (ASR) system usually consists of an acoustic model, a lexicon and a language model that are built and optimized separately. During decoding, the acoustic model, the lexicon and the language model are compiled into a weighted finite-state transducer. In contrast, an ASR system based on a sequence-to-sequence model can directly decode the input speech into text by fully using a single neural network, without a finite state transducer or a lexicon. Therefore, sequence-to-sequence models are becoming increasingly popular in automatic speech recognition (ASR). Comparing to conventional hybrid systems with separate acoustic models, lexicons and language models, a sequence-to-sequence model can be trained in a much simpler way. In addition, it allows joint optimization of all the components and a more compact model. 

There have been a variety of sequence-to-sequence models such as "Listen, Attend and Spell (LAS)" model~\cite{LAS} and recurrent neural network transducer (RNN-T)~\cite{graves2012sequence, graves2013speech, he2019streaming}. Generally, a sequence-to-sequence model processes an input sequence with an encoder recurrent neural network to produce a sequence of hidden states. A decoder RNN is then used to autoregressively produce the output sequence. Unlike conventional ASR, no frame-level alignment is needed during training. Recent results show that the LAS model, which incorporates an attention mechanism between the encoder and the decoder,  has achieved comparable performance with conventional hybrid systems \cite{LAS, spec}.  However, because LAS attends to all hidden states at every output timestep, it must wait until the whole input sequence has been processed before producing any output. Thus it is inapplicable to online speech recognition. Although a RNN-T model can perform streaming recognition, its performance still lags behind that of a large hybrid model~\cite{pundak2016lower}.

In order to make the LAS model suitable for online decoding, the monotonic attention is proposed \cite{monotonic}. However, the monotonic constraints also limits the expressivity of the model. Therefore, Chiu and Raffel proposed to use the monotonic chunk-wise attention (MoChA) in \cite{MoChA}. Unfortunately, experiments show that there are still a gap between MoChA and a vanilla LAS model. In this paper, we aim to improve the MoChA based streaming model, especially for online large vocabulary continuous speech recognition (LVCSR).

Specifically, in this work, we first propose a weight-sharing multi-head MoChA (MTH-MoChA) mechanism to improve the model expressivity. The proposed MTH-MoChA model incorporates a multi-head monotonic attention mechanism. In addition, the parameters for different attention heads are shared to improve the model robustness. A variety of  optimization training strategies are also used to get more improvement on top of the proposed model structure.  Firstly, to reduce the computation complexity, we use pooling between LSTMs~\cite{dong2018extending} to reduce the size of  the sequence. Secondly, SpecAugment~\cite{spec} and label smoothing~\cite{LS} are also adopted during model training. Finally, discriminative minimum word error rate (MWER)~\cite{MWER} training, which is more closely related to word error rate, is applied to get further improvement. The effectiveness of the above structure and training strategies is demonstrated on both an open-source corpus with limited training data and an in-car dataset with 18000-hour of transcribed speech. In the experiments with 18000-hour in-car speech assistant data set, the MTH-MoChA model obtains 7.28\% CER, which is much better than a chain model trained with the lattice free maximum mutual information \cite{LF-MMI} criterion.

The rest of the paper is organized as follows. Section 2 describes the details of our proposed methods and training strategies adopted in this paper. Experiments and results are presented in Section 3, followed by a conclusion in Section 4. 

\section{method details}
\label{sec:format}
In this section, we first briefly introduce the MoChA model. The proposed MTH-MoChA, as well as various optimization training strategies are then described in details. 

\subsection{Monotonic Chunk-wise Attention(MoChA)}
Monotonic attention~\cite{monotonic} explicitly enforces a hard monotonic input-output alignment. Monotonic chunkwise attention (MoChA) \cite{MoChA} relaxes the hard monotonicity constraint and increases the flexibility of the model. MoChA splits the input sequence into small chunks over which the attention is computed. Specifically, given the decoder's state $s_{i-1}$ at step $i-1$ and a sequence of the encoder's hidden states $\{h_1, ..., h_T\}$, the ``energy" $e_{ij}$ used to calculate the attention context is defined as  

\begin{equation}
e_{i,j}=\text{MonotonicEnergy}(s_{i-1},h_j)
\end{equation}
and  
\begin{equation}\label{eq2}
\begin{split}
   &\text{MonotonicEnergy}(s_{i-1},h_j)= \\
   &g\frac{v^T}{\|v\|}\tanh{(W_ss_{i-1}+W_hh_j+b)} + r
\end{split}
\end{equation}
where $W_s \in R^{d \times \text{dim}(s_{i-1})}$, $W_h \in R^{d \times \text{dim}(h_{j})} $, $v \in R^d$, $b \in R^d$, $g \in R$ and $r\in R$ are learnable parameters and $d$ is a hyperparameter (i.e. the hidden dimensionality of the energy function).  

Then, the energy scalars $e_{ij}$ for $j=t_{i-1}, t_{i-1}+1, ..., T$, where $t_i$ is the index of the hidden states chosen at output timestep $i$,  are passed to a logistic sigmoid function $\sigma(.)$ to produce the ``selection probabilities’’  $p_{ij}$:
\begin{equation}
p_{i,j}=\sigma(e_{i,j}+\epsilon),\ \ \ \ \epsilon\sim N(0,1)
\end{equation}
In the above equation, the common approach of adding zero-mean, unit-variance Gaussian noise to the logistic sigmoid function’s activations causes the model to produce effectively binary $p_{ij}$. During decoding, $p_{ij}$ is then used to select the hidden state $h_j$ that the context $c_i$ attends to. The model is trained with respect to the expected value of $c_i$. 

In monotonic attention \cite{monotonic}, the decoder can only attend to a single entry in the hidden states at each timestep. In contrast, MoChA \cite{MoChA} allows the model to perform soft attention over a length-$w$ window of hidden states preceding and including $t_i$:  
\begin{equation}\label{MoChA}
    \begin{split}
        v &= t_i -w +1 \\
        u_{i,k} &= \text{ChunkEnergy}(s_{i-1}, h_k), k \in \{v, v+1, …, t_i\} \\
        c_i &= \sum_{k=v}^{t_i} \text{softmax}(u_{i,k}) h_k
    \end{split}
\end{equation}
Similar to monotonic attention, MoChA can be trained using the expected value of $c_i$. 

\subsection{Multi-head MoChA}
The main difference between MoChA and the proposed multi-head MoChA (MTH-MoChA) lies in the monotonic energy calculation. As can be seen in Eq. (\ref{eq2}), $e_{ij}$ represents the correlation between $s_{i-1}$ and $h_j$. However, the relationship between two high dimensional vectors are very complex. Eq. (\ref{eq2}) is not able to give sufficient information between them. Therefore, we propose to split $s_{i-1}$ and $h_j$ into $K$ heads and calculate the energies between them in order to recover the complex dependency between $s_{i-1}$ and $h_j$. Formally, we have

\begin{equation}
h_j=[h_j^1, h_j^2, ..., h_j^K]
\end{equation}
\begin{equation}
s_{i-1}=[s_{i-1}^1, s_{i-1}^2, ..., s_{i-1}^K]
\end{equation}

For the $k$'th head, the monotonic energy $e_{ij}^k$ can be calculated as 
\begin{equation}
e_{ij}^k=g\frac{v^T}{\|v\|}\tanh(W_ss_{i-1}^{(k)}+W_hh_j^{(k)}+b)+r
\end{equation}
where $W_s \in R^{d \times \text{dim}(s_{i-1}^k)}$, $W_h \in R^{d \times \text{dim}(h_{j}^k)} $, $v \in R^d$, $b \in R^d$, $g \in R$ and $r\in R$ are learnable parameters shared among different heads. On the one hand, sharing weights among different heads can reduce the number of  parameters. On the other hand, it also helps share information among different heads. Because each head calculates a different attention context vector, the proposed MTH-MoChA method can utilize more context information than MoChA. 

Finally, we can obtain a context vector $c_i^k$ as that in MoChA for $K$ different heads. The final context vector $c_i$ is then obtained by averaging the $K$ vectors:
\begin{equation}
c_i=\frac{1}{K}\sum_k{c_i^k}
\end{equation}

\subsection{Optimization strategies}
\label{subsec:optimization}

As will be seen in our experiments, we use various optimization training strateties to further improve the model.

\subsubsection{SpecAugment}

SpecAugment \cite{spec} is a simple but powerful data augmentation method for speech recognition. Attention-based sequence-to-sequence model such as LAS and Transformer is prone to over-fitting. Augmentation helps convert an over-fitting problem into an under-fitting problem. After using SpecAugment, model performance improves greatly.

Generally, the SpecAugment policy consists of time warping, time masking, and frequency masking operation. Despite the warping step, which is less important than other operations, SpecAugment can be regarded as a special block-based dropout layer.

To perform time masking, firstly the length of block $t$ is generated randomly form $[0,T)$, where $T$ is the maximum time masking size. The starting position $t_0$ is then chosen from $[0,\tau-t)$, where $\tau$ is the length of the input sequence. Finally the data between $[{t_0,t}_0+t)$ is dropped. A similar operation is applied in the frequency domain for frequency masking.

\subsubsection{Pooling between LSTMs}
In ASR task, the length of the input feature sequence is much larger than that of the output word sequence. To reduce the length of the input sequence and improve caculating efficiency, maximum pooling between LSTM outputs along time \cite{Aligner} is adopted. Pooling between LSTMs can significantly reduce the computational complexity, as well as improving model's performance because it can remove redundant information from the hidden sequences.

\subsubsection{Minimum Word Error Rate (MWER) Training}
Although Cross Entropy (CE) training is effective, it is not closely related to the optimization metric that we really care about, called word error rate. In the MWER training criterion, the target is to minimize the expected number of word errors directly. The loss function of MWER is shown as follow:
\begin{equation}
L_{MWER}=E_{P(y|x)}[W(y,y^*)]+\lambda L_{CE}
\end{equation}
The loss is interpolated with typical cross-entropy loss to stabilize training \cite{MWER}. When using character as modeling units, MWER is the same as minimum character error rate, or MCER.

\section{Experimental setup and results}
\label{sec:majhead}
\subsection{Data sets}\label{datasets}

We conducted experiments on three Mandarin data sets with different recording settings and sizes to evaluate the effectiveness of the proposed MTH-MoChA model as well as other training strategies. Table~\ref{tab1} summaries the key information about the three data sets. 

\textit{AISHELL-1} is an open-source corpus recorded using high-fidelity microphones in quiet environments such as normal living rooms and recording studios etc. The training set, a total of about 150 hours, contains 120098 utterances from 340 speakers. The text contents are chosen from 11 domains, including finance, science, sports and so on. We used the standard development and testing data sets from \textit{AISHELL-1} to tune and test all the models. 

\textit{InCar2000} and \textit{InCar18000} data sets, with about 2000 and 18000 hours respectively, are collected from Tencent’s in-car speech assistant products.  The speech contents include enquiries,  navigations and conversations. The average number of  characters of each utterance is less than that of the \textit{AISHELL-1} data set. All the data are anonymized and hand-transcribed. The development and testing data sets contain 4998 and 4271 recordings respectively. 

\begin{table}[]
\centering
\caption{Details about the \textit{AISHELL-1}, \textit{InCar2000} and \textit{InCar18000} data sets.}
\label{tab1}
\scalebox{0.85}{
    \begin{tabular}{|c|c|c|c|}
    \hline
    \textbf{data set}                                                         & AISHELL-1                                                                     & InCar2000        & InCar18000       \\ \hline
    \textbf{size(hours)}                                                      & 178                                                                           & 2000       & 18000      \\ \hline
    \textbf{avg. duration}                                                    & \multicolumn{3}{c|}{3$\sim$4s}                                                                                      \\ \hline
    \textbf{avg. characters}                                                  & 14                                                                             & \multicolumn{2}{c|}{6}             \\ \hline
    \textbf{\begin{tabular}[c]{@{}c@{}}recording \\ environment\end{tabular}} & \begin{tabular}[c]{@{}c@{}}quiet living rooms, \\ recording studios etc.\end{tabular} & \multicolumn{2}{c|}{in cars}        \\ \hline
    \textbf{\begin{tabular}[c]{@{}c@{}}text \\ transactions\end{tabular}}     & \begin{tabular}[c]{@{}c@{}}news with various \\ topics\end{tabular}           & \multicolumn{2}{c|}{speech enquiries} \\ \hline
    \end{tabular}
}
\end{table}

\subsection{Model details}

We followed the model configurations as described in~\cite{librispeechLAS}. The encoder consists of 2 convolutional layers with 32-channel $3\times 3$ kernel with a stride of 2, followed by 4 unidirectional LSTM layers with 1024 units. Two-layer CNNs with a stride of 2 for each layer result in a sub-sampling factor of 4 in time. Batch normalization is followed after each CNN and LSTM layer. As to the pooling operation of LSTM, a pooling layer is inserted after the 2nd and the 4th LSTM layer of the encoder to perform max-pooling with width 2. 

The decoder consists of 2 unidirectional LSTM layers with 1024 units. The implementation of the attention follows that in~\cite{weiss2017sequence}. The attention context vector is fed into every layer of the decoder. For MoChA, we use a chunk wise $w=2$ and hidden dimension 1024. For MTH-MoChA, 4 heads are used for all experiments, thus the hidden dimension of each head is 256.

To avoid model over-fitting, label smoothing \cite{LS} with 0.1 uncertainty probability is used. As for SpecAugment, no time-warping is used. The maximum masking block sizes for frequency and time masking are 27 and 40 respectively. In addition, the time masking block size is set to no more than 20\% of the total sequence length.

80-dimensional log-mel features, computed every 10ms with a 25ms window are used as the input. Since there are some English words in the in-car data sets (e.g. English song names in the enquiries), both Chinese and English characters are adopted as modeling units. All the experiments are conducted with the Lingvo \cite{lingvo} toolkit.

We also conducted experiments using state-of-the-art hybrid models on the in-car data sets for comparisons.  TDNN-OPGRU models were trained with the LF-MMI criterion by using Kaldi. We followed the model structure and training scripts described in\cite{OPGRU}. A 18G 5-gram language model was used for TDNN-OPGRU models. However, no language model was used to rescore the output of sequence-to-sequence models. 
 
\subsection{Results on Aishell-1}

We first conducted experiment on \textit{AISHELL-1}. Table~\ref{tab2} gives results of the baseline LAS model, the vanilla MoChA model and the proposed MTH-MoChA model. Results on this data set reported in other work are also given. In table~\ref{tab2}, label smoothing and SpecAugment are used for all end-to-end models. The vanilla non-streaming LAS model obtains 7.51\% character error rate (CER) on the development set and 9.29\% CER on the test set. The performance of the online MoChA model is comparable with that of the LAS model. 

The proposed MTH-MoChA performs slightly better than MoChA on this data set. In addition, the optimization training strategies described in subsection 2.3 significantly improve the model performance. The best sequence-to-sequence model achieves 7.68\% CER on the test set, which is comparable to a strong hybrid model reported in \cite{Aishell-1}. 

\begin{table}[]
\caption{Character error rate (CER) on \textit{AISHELL-1}. Label smoothing and SpecAugment are used for all attention-based end-to-end models. Notice that Transformer\cite{compare} is an offline model.}
\centering
\label{tab2}
\begin{tabular}{ccc}
\hline
\multirow{2}{*}{model} & \multicolumn{2}{c}{CER(\%)} \\ \cline{2-3} 
                       & Dev        & Test        \\ \hline
LFMMI\cite{Aishell-1}               & 6.44   &   7.62    \\
LDS-REG  \cite{siningAdversarial}   & 9.43   &  10.56    \\
LST+RNNLM\cite{LST}                 & 8.30   &      -    \\ 
ESPnet RNN\cite{compare}            & 6.80   &   8.00    \\
ESPnet Transformer\cite{compare}    & 6.00   &   6.70    \\ \hline
LAS                    & 7.51                &   9.29    \\ \hline
MoChA                  & 7.53                &   8.96    \\ \hline
MTH-MoChA              & 7.22                &   8.73    \\
+Pooling+MWER                 & 6.48      &   7.68    \\ \hline
\end{tabular}
\end{table}

\subsection{Results on both \textit{InCar2000} and \textit{InCar18000} corpora}

\begin{table}[]
\label{tab3}
\caption{Experimental results (CER) on our in-car corpora \textit{InCar2000} and \textit{InCar18000}.}
\centering
\begin{tabular}{ccccc}
\hline
\multirow{2}{*}{model} & \multicolumn{2}{c}{InCar2000} & \multicolumn{2}{c}{InCar18000} \\ \cline{2-5} 
                       & Dev        & Test      & Dev       & Test       \\ \hline
TDNN-OPGRU             & 9.25       & 12.26     & 7.40      & 7.94       \\ \hline
LAS                    & 9.55       & 14.72     & -         & -          \\ \hline
MoChA                  & 13.91      & 19.52     & -         & -          \\ \hline
MTH-MoChA              & 10.42      & 15.30     & 7.71      & 8.10       \\
+Pooling+MWER                  & 9.15       & 13.90     & 6.78      & 7.28       \\ \hline
\end{tabular}
\end{table}

To further verify the effectiveness of the MTH-MoChA model and other  optimization training strategies for online attention model, we conduct experiments on another two large data sets, \textit{InCar2000} and \textit{InCar18000}. Table~\ref{tab3} shows results on these two data sets. TDNN-OPGRU is a hybrid model trained with SpecAugment and LFMMI. From the result on \textit{InCar2000}, we can find that MoChA performs much worse than the vanilla LAS model. The reason for this may be that most recordings of the in-car corpus are speech enquiries and the average number of characters inside a recording is much less, meaning that the average duration of utterance segments is shorter. Compared with LAS, MoChA cannot use the whole audio segment to calculate the attention context. Therefore, the available information for attention context calculation is much less, especially for short recordings. 

For \textit{InCar2000} data set, MTH-MoChA obtains 10.42\% and 15.30\% CER the development and test sets respectively. It improves MoChA by 21.62\% relatively on the test data. When combining with the other optimization training strategies, we can get another 9.09\% relative reduction in CER. We did not train the LAS and MoChA models on the \textit{InCar18000} data set since it takes a long time. As can be seen from Tabel \ref{tab3}, the proposed MTH-MoChA model significantly outperforms the TDNN-OPGRU model trained with the LFMMI criterion on the \textit{InCar18000} data set.

\section{Conclusion}
\label{sec:page}
In this paper, we propose multi-head monotonic chunkwise attention (MTH-MoChA) for online large vocabulary speech recognition. MTH-MoChA splits the input sequence into small chunks and computes the multi-head attention context over the chunks. Our experiments show that MTH-MoChA outperforms the original MoChA model. In addition, other optimization training strategies are also necessary to push the performance of MTH-MoChA to state-of-the-art. When a large amount of training data is available, MTH-MoChA, without using any additional language model,  significantly outperforms the best conventional hybrid system. 

In the future, we would like to investigate other training strategies such as schedule sampling (SS) \cite{LAS}, focal loss (FL) \cite{kuaishou}, CTC joint training \cite{compare} and language model (LM) rescoring \cite{LAS} to further improve the performance of the proposed model. By applying these tricks, the online attention model is very likely to get further improvements.
\bibliographystyle{IEEEbib}
\bibliography{Template}

\end{document}